\def\year{2019}\relax
\documentclass[letterpaper]{article} 
\usepackage{aaai19}  
\usepackage{times}  
\usepackage{helvet}  
\usepackage{courier}  
\usepackage{url}  
\usepackage{graphicx}  
\usepackage{makecell}
\usepackage{booktabs}
\usepackage{multirow}
\usepackage{subfigure}
\usepackage{amsmath}

\newcommand{\secref}[1]{Section \ref{#1}}
\newcommand{\figref}[1]{Figure \ref{#1}}

\newcommand{\tabref}[1]{Table \ref{#1}}
\newcommand{\bi}[1]{\textbf{\textit{#1}}}

\newcommand{\lstm}{\operatornamewithlimits{LSTM}}
\newcommand{\bilstm}{\operatornamewithlimits{BiLSTM}}

\newcommand{\seg}{\operatornamewithlimits{seg\_tag}}
\newcommand{\ner}{\operatornamewithlimits{ne\_tag}}
\newcommand{\slot}{\operatornamewithlimits{slot\_fill}}
\newcommand{\batch}{\operatornamewithlimits{Batch}}

\begin{document}
%
\title{Deep Cascade Multi-task Learning for Slot Filling in Online Shopping Assistant}

\author{
	Yu Gong,\textsuperscript{\rm 1}\footnotemark[1]
	Xusheng Luo,\textsuperscript{\rm 1}\footnotemark[1]
	Yu Zhu,\textsuperscript{\rm 1}
	Wenwu Ou,\textsuperscript{\rm 1}
	Zhao Li,\textsuperscript{\rm 1}
	Muhua Zhu,\textsuperscript{\rm 1}\\
	\bf{\Large{Kenny Q. Zhu,\textsuperscript{\rm 2}
	Lu Duan,\textsuperscript{\rm 3}
	Xi Chen\textsuperscript{\rm 1}}} \\
	\textsuperscript{\rm 1}Search Algorithm Team, Alibaba Group \\
	\textsuperscript{\rm 2}Shanghai Jiao Tong University \\
	\textsuperscript{\rm 3}Artificial Intelligence Department, Zhejiang Cainiao Supply Chain Management Co.\\
	\{gongyu.gy, lxs140564, zy143829, lizhao.lz, muhua.zmh\}@alibaba-inc.com, \\
	\{santong.oww, gongda.cx\}@taobao.com, kzhu@cs.sjtu.edu.cn, duanlu.dl@cainiao.com
}

\maketitle

\begin{abstract}
Slot filling is a critical task in natural language understanding
(NLU) for dialog systems.
State-of-the-art approaches treat it as a sequence labeling problem
and adopt such models as BiLSTM-CRF.
While these models work relatively well on standard 
benchmark datasets, they face challenges in the context of
E-commerce where the slot labels are more informative
and carry richer expressions.
In this work, inspired by the unique structure of E-commerce knowledge base,
we propose a novel multi-task model
with cascade and residual connections, which 
jointly learns segment tagging, named entity tagging and slot filling.
Experiments show the effectiveness of the proposed
cascade and residual structures. 
Our model has a 14.6\% advantage in F1
score over the strong baseline methods on a new Chinese E-commerce
shopping assistant dataset, while achieving competitive accuracies on
a standard dataset.  
Furthermore, online test deployed on such dominant E-commerce platform
shows 130\% improvement on accuracy of understanding user utterances.
Our model has already gone into production in the E-commerce platform.
\end{abstract}

\renewcommand{\thefootnote}{\fnsymbol{footnote}}
\footnotetext[1]{Equal contribution.}
\renewcommand{\thefootnote}{\arabic{footnote}}

\section{Introduction}
\label{sec:intro}

An intelligent online shopping assistant
offers services such as pre-sale and after-sale inquiries, 
product recommendations, and user complaints processing,
all of which seek to give the customers better shopping experience.
The core of such assistant is a task-oriented dialog system which 
has the ability to understand natural language utterances
from a user and then give natural language responses \cite{yan2017building}.
Natural Language Understanding (NLU), which aims to 
interpret the semantic meanings conveyed by input utterances, 
is a main component in task-oriented dialog systems. 
Slot filling, a sub-problem of NLU, extracts semantic constituents
by using the words of input utterance to fill in pre-defined slots in 
a semantic frame \cite{mesnil2015using}.

\begin{figure}[th]
	\centering
	\includegraphics[angle=0, width=0.85\columnwidth]{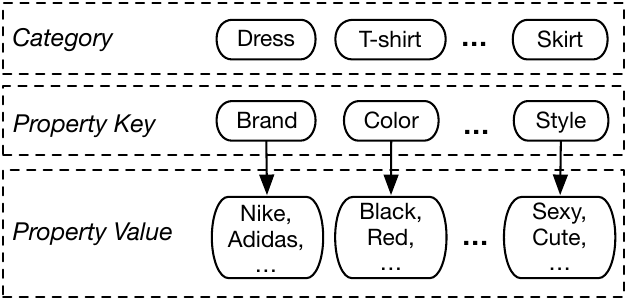}
	\caption{Structure of E-commerce knowledge-base.}
	\label{fig:kg}
\end{figure}

In the case of E-commerce shopping, 
there are three named entity types: 
\emph{Category}, \emph{Property Key} and \emph{Property Value}, according to typical E-commerce knowledge base such as the one in \figref{fig:kg}.
We show a real example in \tabref{tab:slot-filling-demo}
with In/Out/Begin (\textbf{I/O/B}) scheme.
In the named entity level,
``dress'' is a Category (\textbf{CG}),
while ``brand'' is labeled as Property Key (\textbf{PK}),
which is the name of one product property.
``Nike'' and ``black'' are labeled as Property Value (\textbf{PV}) since they 
are concrete property values.
However, merely labeling as Property Value is not sufficient as
the shopping assistant needs more fine-grained semantics.
Therefore, in the Slot Filling level, 
we further label ``Nike'' as Brand Property (\textbf{Brand}), 
and ``black'' as Color Property (\textbf{Color}).
In \tabref{tab:slot-filling-demo}, \textbf{B-CG} refers to 
Begin-Category (the meaning of other labels can also be inferred). 
In the meantime, other words in the example utterance that carry no semantic meaning are assigned \textbf{O} label.
\begin{table*}[h]
	\centering
	\scriptsize
	\begin{tabular}{c|c|c|c|c|c|c|c|c|c|c|c|c|c}
		\toprule
		\multirow{2}{*}{Utterance} & \includegraphics[height=1.8\fontcharht\font`\B]{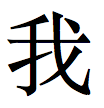} & \includegraphics[height=1.8\fontcharht\font`\B]{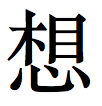} & \includegraphics[height=1.8\fontcharht\font`\B]{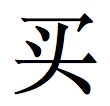} & \includegraphics[height=1.8\fontcharht\font`\B]{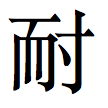} & 
		\includegraphics[height=1.8\fontcharht\font`\B]{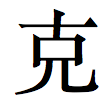} & \includegraphics[height=1.8\fontcharht\font`\B]{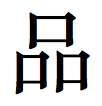} & \includegraphics[height=1.8\fontcharht\font`\B]{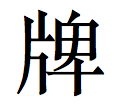} & 
		\includegraphics[height=1.8\fontcharht\font`\B]{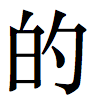} & 
		\includegraphics[height=1.8\fontcharht\font`\B]{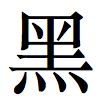} & \includegraphics[height=1.8\fontcharht\font`\B]{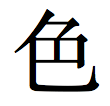} & \includegraphics[height=1.8\fontcharht\font`\B]{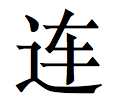} & \includegraphics[height=1.8\fontcharht\font`\B]{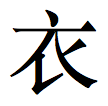} & \includegraphics[height=1.8\fontcharht\font`\B]{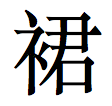} \\
		\cmidrule{2-14}
		& I & want & buy & \multicolumn{2}{c|}{Nike} & \multicolumn{2}{c|}{brand} & $\backslash$  & \multicolumn{2}{c|}{black} & \multicolumn{3}{c}{dress} \\
		\midrule
		Slot Label & \textbf{O} & \textbf{O} & \textbf{O} & \textbf{B-Brand} & \textbf{I-Brand} & \textbf{B-PK} & \textbf{I-PK} & \textbf{O} & \textbf{B-Color} & \textbf{I-Color} & \textbf{B-CG} & \textbf{I-CG} & \textbf{I-CG} \\
		\midrule
		Named Entity Label & \textbf{O} & \textbf{O} & \textbf{O} & \textbf{B-PV} & \textbf{I-PV} & \textbf{B-PK} & \textbf{I-PK} & \textbf{O} & \textbf{B-PV} & \textbf{I-PV} & \textbf{B-CG} & \textbf{I-CG} & \textbf{I-CG} \\
		\midrule
		Segment Label & \textbf{O} & \textbf{O} & \textbf{O} & \textbf{B} & \textbf{I} & \textbf{B} & \textbf{I} & \textbf{O} & \textbf{B} & \textbf{I} & \textbf{B} & \textbf{I} & \textbf{I} \\
		\bottomrule
	\end{tabular}
	\caption{A real example of slot filling in online shopping scenario.}
	\label{tab:slot-filling-demo}
\end{table*}

Traditionally, slot filling problem  can be regarded as a sequence 
labeling task,
which assigns an appropriate semantic label to each word in
the given input utterance.
State-of-the-art sequence labeling models are typically based on 
BiLSTM-CRF \cite{huang2015bidirectional,reimers2017optimal}
and evaluated on a commonly used standard dataset ATIS \cite{price1990evaluation} in the slot filling area.
This dataset is about airline travel in the United States.
However, the vocabulary size of ATIS is small (only 572)
and slot labels are not diverse enough (mostly related to only time and location)
since airline travel is a relatively small and specific domain,
such that recent deep learning models can achieve very high F1 scores
(nearly 0.96). 
Recently, a detailed quantitative and qualitative study of this dataset comes to the same conclusion that 
slot filling models should be tested on a much more real and complex dataset \cite{bechet2018atis}. 

In this paper, we try to tackle a real-world slot filling problem 
for one of the largest E-commerce platform in China.
The semantic slots are much more diverse and informative than ATIS.
For example, to describe different properties of a product for 
the purpose of utterance understanding,
we define large amount of informative slot labels such as color, brand, 
style, season, gender and so on.
In contrast, most semantic labels of ATIS are related to only time and location.
Furthermore, the Chinese language used for e-commerce is more complex 
and the semantically rich expressions make it harder to understand.
Whereas in ATIS, 
expression can be simpler, and most expressions are standard locations or time.
Thus, large scale semantic slots and more complex expressions bring problem such as data sparsity.
Traditional end-to-end sequence labeling model may not be able to handle it.

Besides, Chinese language, like many other Asian languages, is not
word segmented by nature, and word segmentation is a difficult
first step in many NLP tasks.
Without proper word segmentation, sequence labeling becomes very challenging
as the errors from segmentation will propagate.
On the other hand, more than 97\% of the chunks in ATIS data 
have only one or two words,
in which segmentation (or chunking) is not a serious problem.
Due to these reasons,
if we simply apply basic sequence labeling models,
which can be regarded as an end-to-end method,
the sentences may not be segmented correctly in the first place.
Then the errors will propagate and the resulting slot labels will be incorrect.

In this paper, we propose to employ multi-task sequence labeling model
to tackle slot filling in a novel Chinese E-commerce dialog system.
Inspired by the natural structure of E-commerce knowledge base shown in \figref{fig:kg},
we extract two additional lower-level tasks from the slot filling task: 
{\em named entity tagging} and {\em segment tagging}.
Example labels of these two tasks 
are shown in the bottom two rows of \tabref{tab:slot-filling-demo}.
Segment tagging and named entity tagging can be regarded as 
syntactic labeling, 
while slot filling is more like semantic labeling.
With the help of information sharing ability of multi-task learning,
once we learn the information of syntactic structure of an input sentence,
filling the semantic labels becomes much easier.
Compared to directly attacking slot filling,
these two low-level tasks are much easier to solve
due to fewer labels.
To this end, we propose a Deep Cascade Multi-task Learning model,
and co-train three tasks in the same framework
with a goal of optimizing the target slot filling task.

The contributions of this paper are summarized below:
\begin{itemize}
	\item To the best of our knowledge, this is the first piece of work focusing on slot filling in E-commerce. 
	We propose a novel deep multi-task 
	sequence labeling model (DCMTL) with cascading and 
	residual connection to solve it (\secref{sec:dcmtl}).
	\item We develop a Chinese E-commerce shopping assistant dataset 
	ECSA (\secref{sec:data}), which is much bigger and different from 
	the common ATIS dataset, and would be a valuable contribution to 
	dialog system research.  
	\item We evaluate DCMTL in both offline and online settings.
	Offline results show the model outperforms 
	several strong baseline methods by a substantial margin of $14.6\%$ on $F1$ score (\secref{sec:eval}).
	Online testing deployed on the mentioned E-commerce platform shows that 
	slot filling results returned by our model achieve 130\% improvement 
	on accuracy which significantly benefits to the understanding of users' utterances (\secref{sec:case_study}).
	Our model has already gone production in the platform.
\end{itemize}
	
\section{Approach}
\label{sec:model}
\begin{figure}[th]
	\centering
	\includegraphics[angle=0, width=1.0\columnwidth]{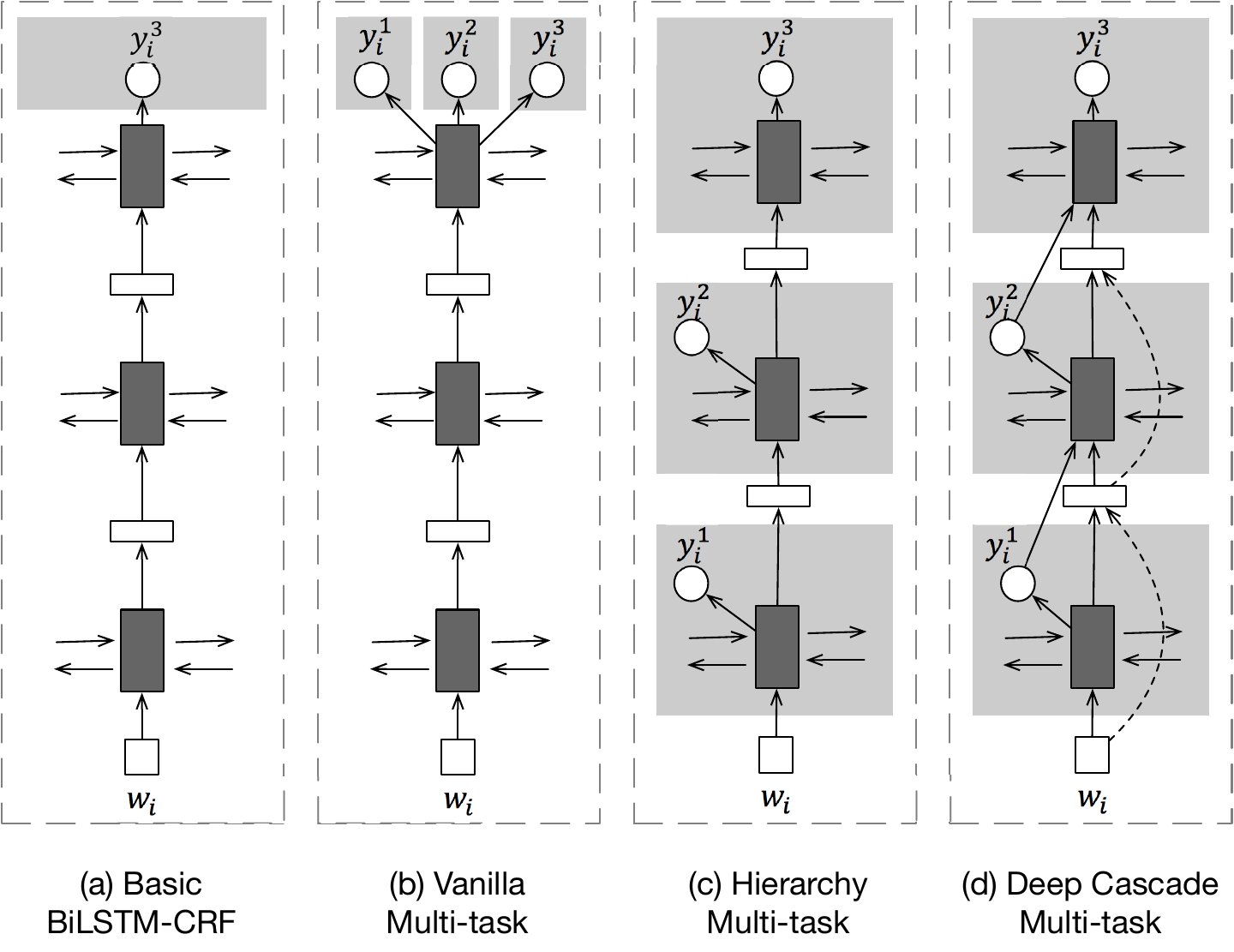}
	\caption{Sequential models for slot filling task.}
	\label{fig:model}
\end{figure}

In this section we describe our approach in detail.
\figref{fig:model} gives an overview of the proposed architectures.
First we introduce the most common and popular BiLSTM-CRF
model (\figref{fig:model}(a)) for sequence labeling tasks.
Then we move on to multi-task learning perspective 
(\figref{fig:model}(b) and (c)).
Finally we propose our new method, which is called
Deep Cascade Multi-task Learning in \figref{fig:model}(d).

Given an utterance containing a sequence of words
$\textbf{w} = (w_1, w_2, ..., w_T)$,
the goal of our problem
is to find a sequence of slot labels $\hat{\textbf{y}} = (y_1, y_2, ..., y_T)$, 
one for each word in the utterance, such that:
\begin{equation*}
\hat{\textbf{y}} = \mathop{\arg\max}_{\textbf{y}}P(\textbf{y}|\textbf{w}).
\end{equation*}
We use ``word'' in problem and model description,
but ``word'' actually means Chinese char in our problem.
And a ``term'' consists of one or several words.

\subsection{RNN Sequence Labeling}
\label{sec:rnn_sequence_labeling}

\figref{fig:model}(a) 
shows the principle architecture of a BiLSTM-CRF model,
which is the state-of-the-art model for various sequence labeling tasks \cite{huang2015bidirectional,reimers2017optimal}.
BiLSTM-CRF model consists of a BiLSTM layer and a CRF layer. 

BiLSTM (Bidirectional-LSTM) enables the
hidden states to capture both historical and future
context information of the words.
Mathematically, the input of this BiLSTM layer
is a sequence of input vectors,
denoted as $\bi{X}=(\bi{x}_1, \bi{x}_2, ..., \bi{x}_T)$.
The output of BiLSTM layer is a sequence of the hidden
states for each input word, denoted
as $\bi{H}=(\bi{h}_1, \bi{h}_2, ..., \bi{h}_T)$.
Each final hidden state is the concatenation of the forward
$\overrightarrow{\bi{h}_i}$ and backward $\overleftarrow{\bi{h}_i}$ hidden states.
We view BiLSTM as a function $\bilstm(\bi{x}_i)$:
\begin{eqnarray*}
	& \overrightarrow{\bi{h}_i} = \lstm(\bi{x}_i, \overrightarrow{\bi{h}_{i-1}}),
	\overleftarrow{\bi{h}_i} = \lstm(\bi{x}_i, \overleftarrow{\bi{h}_{i+1}}), \\
	& \bilstm(\bi{x}_i) = \bi{h}_i = [\overrightarrow{\bi{h}_i}(\bi{x}_i);\overleftarrow{\bi{h}_i}(\bi{x}_i)].
\end{eqnarray*}
Most of time we stack multiple BiLSTMs to make the model deeper,
in which the output $\bi{h}_i^l$ of layer $l$ becomes the input of layer $l+1$,
e.g. $\bi{h}_i^{l+1}=\bilstm^{l+1}(\bi{h}_i^l)$.

It is always beneficial to consider the correlations
between the current label and neighboring
labels, since there are many syntactical constraints
in natural language sentences.
For example,
\textbf{I-Brand} is never followed by a \textbf{B-Color}.
If we simply feed the above mentioned hidden states
independently to a softmax layer to predict the labels \cite{hakanni-tur2016multidomain},
such constraints are more likely
to be violated. Linear-chain Conditional Random
Field (CRF) \cite{lafferty2001conditional} 
is the most popular way to control the structure
prediction and its basic idea is to use a series
of potential functions to approximate the conditional
probability of the output label sequence
given the input word sequence.

Formally, we take the above sequence of hidden
states $\bi{H} = (\bi{h}_1, \bi{h}_2, ..., \bi{h}_T)$
as input to a CRF layer,
and the output of the CRF is the final prediction label sequence
$\bi{y} = (y_1, y_2, ..., y_T)$,
where $y_i$ is in the set of pre-defined target labels.
We denote $\mathcal{Y}(\bi{H})$ as the set of all possible label sequences.
Then we derive the conditional probability of the output sequence,
given the input hidden state sequence is:
\begin{equation*}
p(\bi{y}|\bi{H};\bi{W}, \bi{b})=\frac{\prod_{i=1}^{T}\varphi(y_{i-1},y_i,\bi{H})}
{\sum_{\bi{y}'\in\mathcal{Y}(\bi{H})}\prod_{i=1}^{T}\varphi(y'_{i-1},y'_i,\bi{H})},
\end{equation*}
where $\varphi(y',y,\bi{H})=\exp(\bi{W}_{y',y}^{T}\bi{H}+\bi{b}_{y',y})$ are potential functions and $\bi{W}_{y',y}^{T}$ and $\bi{b}_{y',y}$ are weight vector and bias of label pair $(y', y)$.
To train the CRF layer, we use the classic maximum
conditional likelihood estimate and gradient ascent.
For a training dataset $\{(\bi{H}_{(i)}, \bi{y}_{(i)})\}$, 
the final log-likelihood is:
\begin{equation*}
L(\bi{W},\bi{b}) = \sum_{i}\log p(\bi{y}_{(i)}|\bi{H}_{(i)};\bi{W},\bi{b}).
\end{equation*}
Finally, the Viterbi algorithm is adopted
to decode the optimal output sequence $\bi{y}^{*}$:
\begin{equation*}
\bi{y}^{*}=\mathop{\arg\max}_{\bi{y}\in\mathcal{Y}(\bi{H})}p(\bi{y}|\bi{H};\bi{W},\bi{b}).
\end{equation*}

\subsection{Multi-task Learning}
The slot labels are large-scaled, informative and diverse in the case of E-commerce,
and the syntactic structure of input Chinese utterance are complicated,
so that the slot filling problem becomes hard to solve.
If we directly train an end-to-end sequential model,
the tagging performance will suffer from data sparsity severely.
When we try to handle slot filling (can be seen as semantic labeling task),
some low-level tasks such as named entity tagging or segment tagging (can be seen as syntactic labeling task)
may first make mistakes.
If the low-level tasks get wrong, so as to the target slot filling task.
That is to say it is easy to make wrong decisions in the low-level tasks,
if we try to fill in all the labels at once.
Then the error will propagate and lead to a bad performance of slot filling, which is our high-level target.

While directly attacking the slot filling task is hard,
low-level tasks with fewer labels are much easier to solve.
Once we know the syntactic structure of a sentence,
filling in semantic labels will become easier accordingly.
Thus, it is reasonable to solve the problem in a multi-task learning framework.
In our problem, 
following the special structure of E-commerce knowledge base (\figref{fig:kg}),
we can devise three individual tasks: slot filling, named entity tagging and segment tagging.
Slot filling is our target task;
named entity tagging is to classify which named entity type (\textbf{PV}/\textbf{PK}/\textbf{CG}) a word is;
and segment tagging is to judge 
whether a word is begin (\textbf{B}), in (\textbf{I}) or out (\textbf{O}) of a trunking.

In a multi-task learning (MTL) setting, 
we have several prediction tasks over the same input sequence,
where each task has its own output vocabulary (a set of task 
specified labels).
Intuitively, the three tasks do share a lot of information.
Consider the example in \tabref{tab:slot-filling-demo} again. 
Knowing the named entity type of 
``\includegraphics[height=1.4\fontcharht\font`\B]{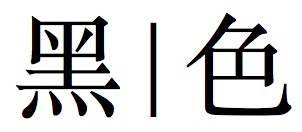}'' being 
\textbf{B-PV}$|$\textbf{I-PV}
can definitely help determine its slot label, 
which is \textbf{B-Color}$|$\textbf{I-Color}.
Similarly, knowing its segment type (\textbf{B}$|$\textbf{I}) 
also helps with both named entity tagging and slot filling.
Thus it is reasonable for these tasks to share parameters
and learn in the same framework cooperatively.

\subsubsection{Vanilla Multi-task Learning}
The general idea of multi-task learning
is to share parameters of encoding part of the network.
As \figref{fig:model}(b) shows,
this is naturally achieved by sharing 
the $k$-layers BiLSTM part of the network across three tasks.
Based on that,
we use a separate CRF decoder for each task $t \in \{seg, ne, slot\}$:
$p(\bi{y}^t|\bi{H}^k;\bi{W}_t, \bi{b}_t)$,
where $\bi{W}_t$ and $\bi{b}_t$ are task-specific parameters.
This encourages the deep BiLSTM network 
to learn a hidden representation $\bi{H}^k$
which benefits all three different tasks.

\subsubsection{Hierarchy Multi-task Learning}
Previous discussion indicates that 
there is a natural order among the different tasks:
slot filling may benefit more from named entity tagging, than the other way around.
This motivates us to employ low-level tasks at lower BiLSTM layers,
while high level tasks are trained at higher layers.
We borrow the idea of involving a hierarchical neural networks structure \cite{peters2018deep,sogaard2016deep}.
As shown in \figref{fig:model}(c), 
instead of decoding all tasks separately at the outermost BiLSTM layer, 
we associate each BiLSTM layer
$l(t)$ with one task $t$.
Then the conditional probabilities of the output sequence for each task are:
\begin{eqnarray*}
	& \seg(\bi{w})=p(\bi{y}^{seg}|\bi{H}^{l(seg)};\bi{W}_{seg},\bi{b}_{seg}), \\
	& \bi{H}^{l(seg)}=\bilstm^{l(seg)}(E(\bi{w})). \\
	& \ner(\bi{w})=p(\bi{y}^{ne}|\bi{H}^{l(ne)};\bi{W}_{ne},\bi{b}_{ne}), \\
	& \bi{H}^{l(ne)}=\bilstm^{l(ne)}(\bi{H}^{l(seg)}).\\
	& \slot(\bi{w})=p(\bi{y}^{slot}|\bi{H}^{l(slot)};\bi{W}_{slot},\bi{b}_{slot}), \\
	& \bi{H}^{l(slot)}=\bilstm^{l(slot)}(\bi{H}^{l(ne)}).
\end{eqnarray*}
Here $\seg$, $\ner$ and $\slot$ represent the tasks of
segment tagging, named entity tagging and slot filling, respectively.
$E(\bi{w})$ is the word embeddings of input sequence $\bi{w}$ and 
$l(seg) < l(ne) < l(slot)$.
We call this model hierarchy multi-task learning,
since some layers are shared by all tasks 
while the others are only related to specific tasks.

\subsection{Deep Cascade Multi-task Learning}
\label{sec:dcmtl}
Hierarchy multi-task learning share parameters among different tasks,
and allow low-level tasks help adjust the result of high-level target task.
It is effective for those tasks which are weakly correlated,
such as POS tagging, syntactic chunking and CCG supertagging \cite{sogaard2016deep}.
However, when it comes to problems where different tasks maintain a 
strict order, in another word, the performance of high-level task 
dramatically depends on low-level tasks,
the hierarchy structure is not compact and effective enough.
Therefore, we propose \emph{cascade} and \emph{residual} connections
to allow high-level tasks to take the tagging results and hidden states 
from low-level tasks as additional input. 
These connections serves as ``shortcuts'' that 
create a more closely coupled and efficient model.
We call it deep cascade multi-task learning, 
and the framework is shown in \figref{fig:model}(d).

\subsubsection{Cascade Connection}
Here we feed the tagging output of the task at lower layer 
e.g. $\seg^*(\bi{w})$ or $\ner^*(\bi{w})$
to the upper BiLSTM layer as its additional input.
Now the hidden states of each task layer become:
\begin{eqnarray*}
	& \bi{H}^{l(seg)}=\bilstm^{l(seg)}(E(\bi{w})), \\
	& \bi{H}^{l(ne)}= \bilstm^{l(ne)}(\bi{W}_{Cas.}^{seg}\cdot \seg^*(\bi{w})+\bi{H}^{l(seg)}), \\
	& \bi{H}^{l(slot)}= \bilstm^{l(slot)}(\bi{W}_{Cas.}^{ne}\cdot \ner^*(\bi{w})+\bi{H}^{l(ne)}),
\end{eqnarray*}
where $\bi{W}_{Cas.}^{seg}$ and $\bi{W}_{Cas.}^{ne}$ are the weight parameters for cascade connection.

At training time, $\seg^*(\bi{w})$ and $\ner^*(\bi{w})$ can be the 
true tagging outputs.
At inference time, we simply take the greedy path of our cascade model 
without doing search, where the model emits the
best $\seg^*(\bi{w})$ and $\ner^*(\bi{w})$ by Viterbi inference algorithm.
Alternatively, one can do beam search \cite{sutskever2014sequence,vinyals2015show} by maintaining a set of $k$ best partial hypotheses at each cascade layer.
However, unlike traditional seq2seq models e.g., in machine translation,
where each inference step is just based on probability of a discrete variable 
(by softmax function), our inference for tagging output is a 
structured probability distribution defined by the CRF output.
Efficient beam search method for this structured cascade model is 
left to our future work.

\subsubsection{Residual Connection}
To encourage the information sharing among different tasks,
we also introduce the residual connection,
where we add the input of a previous layer to the current input:
\begin{eqnarray*}
	& \bi{H}^{l(seg)}=\bilstm^{l(seg)}(\bi{x}^{l(seg)}), \\
	& \bi{x}^{l(seg)}=E(\bi{w}). \\
	& \bi{H}^{l(ne)}= \bilstm^{l(ne)}(\bi{W}_{Cas.}^{seg}\cdot \seg^*(\bi{w})+\bi{x}^{l(ne)}), \\
	& \bi{x}^{l(ne)}= \bi{H}^{l(seg)}+\bi{x}^{l(seg)}. \\
	& \bi{H}^{l(slot)}= \bilstm^{l(slot)}(\bi{W}_{Cas.}^{ne}\cdot \ner^*(\bi{w})+\bi{x}^{l(slot)}), \\
	& \bi{x}^{l(slot)}= \bi{H}^{l(ne)}+\bi{x}^{l(ne)}.
\end{eqnarray*}

Deep residual learning \cite{he2016deep} is introduced to ease the gradient vanish problem for training very deep neural networks.
Here we borrow the idea of cross residual learning method for multi-task visual recognition \cite{jou2016deep}
and believe the residual connection between different layers can benefit our multi-task sequence learning.
We propose \emph{cascade} residual connection instead of \emph{cross} residual connection
because different tasks are connected via cascading in our problem, while they are organized via branching in visual recognition.

\subsection{Training}
\label{sec:training}
For our multi-task setting, 
we define three loss functions (refer to \secref{sec:rnn_sequence_labeling}):
$L_{seg}$, $L_{ne}$ and $L_{slot}$ for tasks of segment tagging, named entity tagging 
and slot filling respectively.
We construct three training set,
$D_{seg}$, $D_{ne}$ and $D_{slot}$,
where each of them (called $D_t$ generically) contains a set of 
input-output sequence pair $(\bi{w}, \bi{y}^t)$.
The input utterance $\bi{w}$ is shared across tasks, but the output $\bi{y}^t$ is task dependent.

For vanilla multi-task learning,
we define a unified loss function $L=\alpha L_{seg}+\beta L_{ner}+(1-\alpha-\beta)L_{slot}$, 
where $\alpha$ and $\beta$ are hyper-parameters.
And we update the model parameters by loss $L$.

As for hierarchy multi-task learning and cascade multi-task learning,
we choose a random task $t \in \{seg, ne, slot\}$ at each training step,
followed by a random training batch $\batch(\bi{w}, \bi{y}^t) \in D_t$.
Then we update the model parameters by back-propagating
the corresponding loss $L_t$.
	
\section{Experiments}
\label{sec:exp}
In this section we first introduce the popular ATIS dataset\footnote{\url{https://github.com/yvchen/JointSLU}},
then describe how we collect our \textbf{E}-\textbf{c}ommerce \textbf{S}hopping \textbf{A}ssistant (ECSA) dataset\footnote{\url{https://github.com/pangolulu/DCMTL}}.
Then we show the implementation details for our model.
Finally we demonstrate the evaluation results on both ATIS and ECSA dataset and give some discussions.
In the following experiments, we call our proposed \textbf{D}eep \textbf{C}ascade \textbf{M}ulti-\textbf{T}ask \textbf{L}earning method as DCMTL for short.

\subsection{Dataset}
\label{sec:data}
\subsubsection{ATIS Dataset}
The ATIS corpus, the most commonly used dataset for slot filling research, contains reservation requests for air travel.
It contains 4,978 training and 893 testing sentences in total, with a vocabulary size of 572 \cite{mesnil2015using}. 
Apart from the ground-truth slot labels, we also generate its 
corresponding segment labels for our multi-task model setting.

\subsubsection{ECSA Dataset}
\label{sec:ECSGA_data}
To create large amounts of gold standard data 
to train our model,
we adopt an unsupervised method to 
automatically tag the input utterances.
All the utterances are extracted from the user input logs 
(either from text or voice) on our online shopping 
assistant system. Besides our E-commerce knowledge-base
is a dictionary consisting of pairs of
word terms and their ground-truth slot labels 
such as ``red-\textbf{Color}'' or ``Nike-\textbf{Brand}''.
Since this resource is created by human beings,
we will use it to create gold standard.
We use a dynamic programming algorithm of max-matching to 
match words in the utterances and then assign each word with 
its slot label in IOB scheme.
We filter utterances whose matching result is
ambiguous and only reserve those that can be 
perfectly matched
(all words can be tagged by only one unique label) 
as our training and testing data.
With the slot labels of each word, we can induce the 
named entity labels and segment labels straightforwardly via the E-commerce knowledge-base.
For we only extract the perfectly matched sentences,
the quality of our ECSA dataset can be guaranteed.
It can be considered as a long-distance supervision method \cite{mintz2009distant}.

To evaluate model's ability to generalize, 
we randomly split the dictionary into three parts.
One part is used to generate testing data and the other two 
to generate training data.
If we don't split the dictionary and use the whole to generate 
both training and testing data,
then the trained model may remember the whole dictionary and 
the results will not reflect the true performance of the models.

This unsupervised approach 
alleviates human annotations, 
and we can produce a large volume of labeled data automatically. 
The following experiments use a dataset of 24,892 training pairs
and 2,723 testing pairs.
Each pair contains an input utterance $\bi{w}$, 
its corresponding gold sequence of slot labels $\bi{y}^{slot}$,
named entity labels $\bi{y}^{ne}$ and segment labels $\bi{y}^{seg}$.
The vocabulary size of ECSA is 1265 (Chinese characters), and the amount of segmented terms can be much larger.
The Out-of-Vocabulary (OOV) rate of ESCA dataset is 85.3\% 
(Meaning 85.3\% of terms in testing data never appear in training data) 
while the OOV rate of ATIS is lower than 1\%.
Apparently slot filling task on ESCA dataset is more challenging. 

\subsection{Implementation Details}
\label{sec:implementation}
For the RNN component in our system,
we use a 3-layers BiLSTM networks for ECSA and 2-layers BiLSTM networks for ATIS (no named entity tagging in this case),
and all LSTM networks come with hidden state size 100.
The input in ECSA is a sequence of Chinese characters rather that words 
since there is no segmentation.
The dimension of embedding layer $E$ 
and BiLSTM network output state (concatenation of the forward and backward LSTM) are set to 200.
We perform a mini-batch log-likelihood loss training with a batch size of 
32 sentences for 10 training epochs.
We use Adam optimizer, and the learning rate is initialized to 0.001.
To prevent the gradient explosion problem for training LSTM networks,
we set gradient clip-norm as 5.

\subsection{Results and Discussions}
\label{sec:eval}

\subsubsection{Evaluation on ATIS}
We compare the ATIS results of our DCMTL model with current published results in \tabref{tab:eval_ATIS}.
We split the methods into two categories:
one is \emph{Sequence Labeling} based method, and the other is \emph{Encoder-Decoder} based method.
Sequence Labeling based method generally adopts a sequential network
(RNN \cite{yao2013recurrent,yao2014spoken,liu2015recurrent,peng2015recurrent,vu2016bi} or 
CNN \cite{xu2013convolutional,vu2016sequential})
and calculate a loss function (such as CRF loss \cite{xu2013convolutional}, 
cross entropy loss \cite{yao2013recurrent,yao2014spoken} or 
ranking loss \cite{vu2016bi}) on top of the network output.
Encoder-Decoder based method, on the other hand, 
usually employs a RNN to encode the whole sentence 
and another RNN to decode the labels \cite{kurata2016leveraging}.
The decoder will attend to the whole encoding sequence with attention mechanism \cite{zhu2017encoder,zhai2017neural}.
Our method follows the Sequence Labeling framework 
and we design a novel multi-task sequence labeling model
which achieve the best performance against the published Sequence Labeling based method (F1+0.22\%)
and compatible result against the best Encoder-Decoder based method (F1-0.03\%).
As we claim in \secref{sec:intro}, 
more than 97\% of chunks in ATIS dataset have only one or two words and there are no named entity labels at all.
These two reasons prevent our proposed DCMTL model from further improving the performance on ATIS dataset.
Thus, we will mainly focus on ECSA dataset, 
which is much larger and more sophisticated,
to prove the effectiveness of our proposed model.

Besides, almost all the methods (including ours) 
reach very high F1 score of around 0.96.
This also makes us wonder
whether it is meaningful enough to continue evaluating on this dataset,
for minor differences in the results may be attributed to data variance
more than the models.
Apparently high performance on ATIS does not 
mean working on real-world application
which contains more informative semantic slot labels and more complicated 
expressions as in the case of online shopping assistant.
\begin{table}[th]
	\centering
	\scriptsize
	\begin{tabular}{c|c}
		\toprule
		Methods & F1 \\
		\midrule
		simple RNN \cite{yao2013recurrent} & 0.9411 \\
		CNN-CRF \cite{xu2013convolutional} & 0.9435 \\
		LSTM \cite{yao2014spoken} & 0.9485 \\
		RNN-SOP \cite{liu2015recurrent} & 0.9489 \\
		Deep LSTM \cite{yao2014spoken} & 0.9508 \\
		RNN-EM \cite{peng2015recurrent} & 0.9525 \\
		Bi-RNN with ranking loss \cite{vu2016bi} & 0.9547 \\
		Sequential CNN  \cite{vu2016sequential} & \underline{0.9561} \\
		\midrule
		Encoder-labeler Deep LSTM \cite{kurata2016leveraging} & 0.9566 \\
		BiLSTM-LSTM (focus) \cite{zhu2017encoder} & 0.9579 \\
		Neural Sequence Chunking \cite{zhai2017neural} & \textbf{0.9586} \\
		\midrule
		DCMTL (Ours) & $\underline{0.9583}^{*}$  \\
		\bottomrule
	\end{tabular}
	\caption{Comparison with published results on the ATIS dataset.}
	\label{tab:eval_ATIS}
\end{table}

\subsubsection{Evaluation on ECSA}
\begin{table}[th]
	\centering
	\scriptsize
	\begin{tabular}{c|ccc}
		\toprule
		Models & Precision & Recall & F1 \\
		\midrule
		Basic BiLSTM-CRF & 0.4330 & 0.4275 & 0.4302 \\
		* Basic BiLSTM-CRF (cond. SEG) & 0.7948 & 0.7953 & 0.7950 \\
		* Basic BiLSTM-CRF (cond. NE) & 0.8985 & 0.8986 & 0.8985 \\
		\midrule
		Vanilla Multi-task & 0.3990 & 0.3941 & 0.3965 \\
		Hierarchy Multi-task & 0.4417 & 0.4494 & 0.4455 \\
		\midrule
		** DCMTL (\textbf{-} cascade) & 0.4654  & 0.4613 & 0.4633  \\
		** DCMTL (\textbf{-} residual) & 0.4923 & 0.4760 & 0.4840  \\
		DCMTL (full) & \textbf{0.5281} & \textbf{0.4941} & \textbf{0.5105} \\
		\bottomrule
	\end{tabular}
	\caption{Results for slot filling task on the ECSA dataset.
		Columns with highlighted boldface are the best performance.
		Rows with * prefix are just results for our case study.
		Rows with ** prefix are results for ablation test.}
	\label{tab:eval_ECSGA}
\end{table}

\begin{figure}[h]
	\centering
	\subfigure[]{\includegraphics[width=0.49\columnwidth]{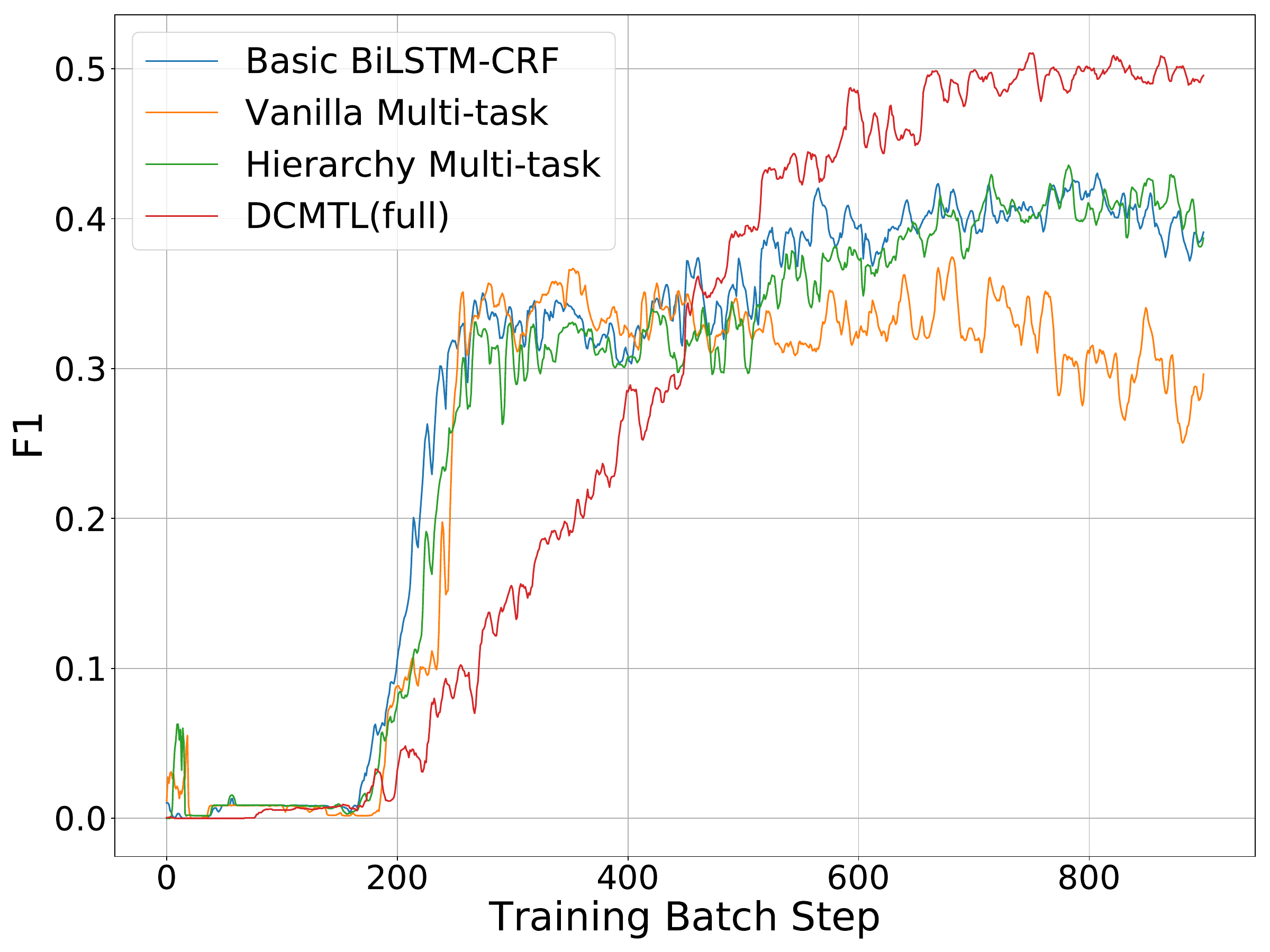}}
	\subfigure[]{\includegraphics[width=0.49\columnwidth]{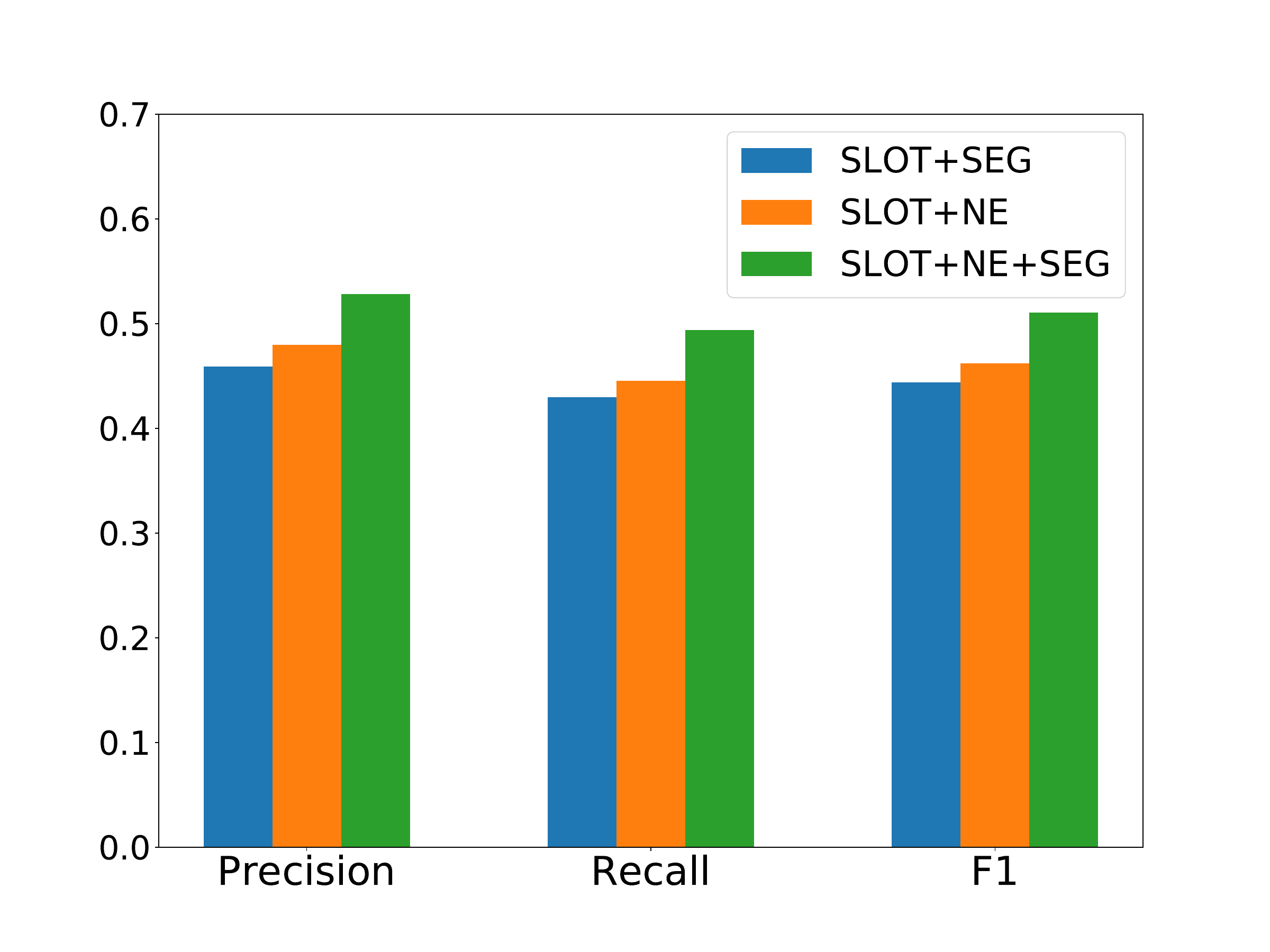}}
	\caption{(a) Learning trends of F1 respectively for different methods.
		(b) Result of different cascade connection types in DCMTL.}
	\label{fig:learning_curve}
\end{figure}

On ECSA dataset,
we evaluate different models including 
Basic BiLSTM-CRF, Vanilla Multi-task, 
Hierarchy Multi-task and Deep Cascade Multi-task
on testing data regarding slot filling as the target task.
We report Precision, Recall and F1 in \tabref{tab:eval_ECSGA}.

The Basic BiLSTM-CRF model achieves an F1 score of 0.43.
To show the usefulness of the lower tasks to slot filling,
we ``cheated'' by using the ground-truth segment type (cond. SEG) or 
named entity type (cond. NE) as the extra features for each word
in the Basic BiLSTM-CRF model.
Row 3 and 4 (with *) in \tabref{tab:eval_ECSGA} show that the slot filling 
performance can be improved by 85\% and 109\% if the correct segment type
or named entity type is pre-known.
It can perfectly verify our claim that low-level syntactic tasks can significantly affect to the slot filling performance.
Of course in practice, the model doesn't know the true values of these types during prediction.

Our further experiments show that DCMTL outperforms the 
baselines on both precision and recall.
DCMTL achieves the best F1 score of 0.5105, 
which improves by a relative margin of 14.6\% 
against the strong baseline method (see \tabref{tab:eval_ECSGA}).
Multi-task models generally perform better than the Basic 
BiLSTM with single-task target.
The exception is the vanilla multi-task setting.
This is mainly because 
vanilla multi-task shares parameters across all the layers,
and these parameters are likely to be disturbed by the interaction of 
three tasks. It is more desirable to let the target task dominate the 
weights at high-level layers.

We further investigate the learning trend of our proposed approach against baseline methods.
\figref{fig:learning_curve}(a) shows the typical learning curves 
of performance measured by F1.
We can observe that our method DCMTL 
performs worse than other baseline methods
for the first $450$ batch steps.
After that, other methods converge quickly 
and DCMTL perform much better after $500$ batch steps
and finally converge to the best F1 score.
We believe that in the beginning,
high-level task in DCMTL is affected more by the noise of low-level tasks comparing to others,
but as the training goes on,
the high-level slot filling task slowly reaps the benefits from low-level tasks.

To make our experiments more solid, 
we implemented two previous best performing models on ATIS dataset: 
Sequential CNN \cite{vu2016sequential} (Sequence Labeling based) and Neural Sequence Chunking \cite{zhai2017neural} (Encoder-Decoder based).
They achieved 0.2877 and 0.4355 F1 scores respectively, while
our DCMTL model scores 0.5105 F1 and outperforms both of them (by \textbf{77\%} and \textbf{17\%} improvements).

\subsubsection{Ablation Test}
Our ``shortcuts'' connections come in two flavors: cascade connection and residual connection.
Multi-task outputs and ``shortcuts'' connections are highly related since without the multi-task framework, there will be no cascade connections.
We go on to show that both multi-task setting and the ``shortcuts'' connections are effective and useful in \tabref{tab:eval_ECSGA},
where F1 score improves from 0.4302 to 0.4455 and 0.5105 respectively.
We also investigate how our model DCMTL performs with or without cascade and residual connections (rows with ** prefix in \tabref{tab:eval_ECSGA}).
F1 score increases from 0.4840 to 0.5105 when residual connection is applied, which verifies its benefit.
If we remove cascade connection from DCMTL,
the model actually degenerates into hierarchy multi-task model with residual connection and performs 0.4633 F1 score.
Thus we can conclude that both connections are helpful for our DCMTL model.
However, the cascade connection, which relies on the multi-task, is more effective than the residual connection.
We can verify it from the fact that DCMTL model without cascade connection performs much worse than without residual connection (0.4633 vs. 0.4840 F1 scores).

\begin{figure}[h]
	\centering
	\subfigure[]{\includegraphics[width=0.49\columnwidth]{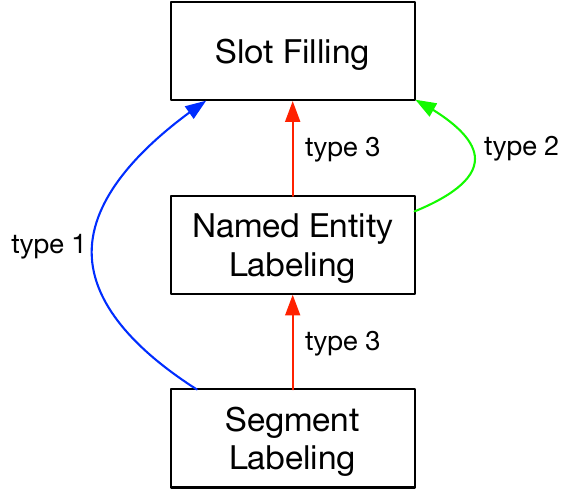}}
	\subfigure[]{\includegraphics[width=0.49\columnwidth]{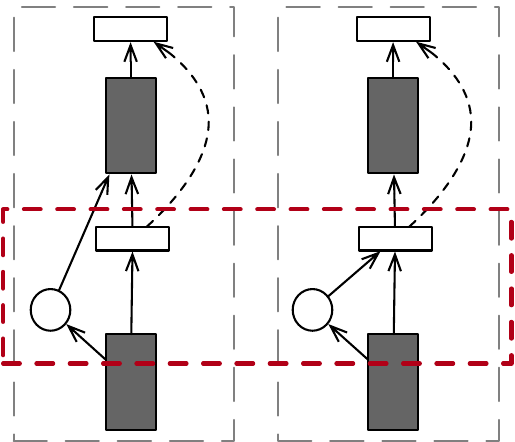}}
	\caption{(a) Three types of cascade connection in our experiment.
		(b) Comparison between hierarchical and skipped cascade connection.}
	\label{fig:cascade_connection_type}
\end{figure}

Furthermore, 
we explore how DCMTL performs with different cascade connection methods.
We compare three different types of cascade connection 
illustrated in \figref{fig:cascade_connection_type}(a):
\begin{enumerate}
	\item[1.]
	Segment labeling skipped to slot filling (SLOT+SEG).
	\item[2.]
	Named entity labeling directly connected to slot filling \\(SLOT+NE).
	\item[3.]
	Segment labeling, named entity labeling and slot filling in sequence (SLOT+NE+SEG).
\end{enumerate}

From \figref{fig:learning_curve}(b),
we find that cascade connection with type 3 
performs the best
and then with type 2,
while cascade method with skipped connection (type 1) performs the worst.
Therefore, we design the networks 
with a cascade connection in a hierarchical fashion
and do not apply skipped connection for the cascade 
inputs (\figref{fig:cascade_connection_type}(b)).
This phenomenon here may also be proved by our ``cheated'' case study above.
Slot filling performance with pre-known named entity type is 
much better than with pre-known segment type
(rows with * in \tabref{tab:eval_ECSGA}).

\subsection{Online Testing}
\label{sec:case_study}
Previous experimental results have proven the advantages of
our proposed DCMTL approach, so we deploy it in a real
world online environment to test its practical performance.

For online A/B testing, we extracted users query log with the slot filling results for one day. There are in total 251,409 unique queries.
We let three persons to manually evaluate whether a query is slotted perfectly with the strategy where the minority obeys the majority.
A query is slotted perfectly means all terms in query are assigned with the correct slot labels.
Our DCMTL model results in 152,178 perfectly slotted queries which is \textbf{60.53\%} accuracy\footnote{We only report accuracy as evaluation metric,
because precision and recall are the same in such case.}.
While the original online max-matching algorithm with E-commerce knowledge base\footnote{As we have showed that our DCMTL model outperforms several strong baselines in the offline evaluation, and the gap between online and offline is minor since our offline dataset also comes from online queries, 
we only deploy DCMTL model online since such evaluation is costly.} 
(more details in \secref{sec:ECSGA_data})
only covers 66,302 perfectly slotted queries with \textbf{26.37\%} accuracy.
Thus, the accuracy of query slot filling in such online shopping assistant 
system is improved by 130\% after deploying DCMTL model.
This demonstrates that our model can effectively
extract the semantic attributes of users query which is extremely helpful E-commerce Shopping Assistant system.
	
\section{Related Work}
\label{sec:related}
There are mainly two lines of research that are related to our work:
slot filling for dialog system and multi-task learning in natural language processing.

\textbf{Slot Filling} is considered a sequence labeling problem
that is traditionally solved by generative models.
such as Hidden Markov Models (HMMs) \cite{wang2005spoken},
hidden vector state model \cite{he2003data}, 
and discriminative models such as 
conditional random fields (CRFs) \cite{raymond2007generative,lafferty2001conditional} 
and Support Vector Machine (SVMs) \cite{kudo2001chunking}.
In recent years,
deep learning approaches have been explored
due to its successful application in many NLP tasks.
Many neural network architectures have been used such as
simple RNNs \cite{yao2013recurrent,mesnil2015using}, 
convolutional neural networks (CNNs) \cite{xu2013convolutional},
LSTMs \cite{yao2014spoken} 
and variations like encoder-decoder \cite{zhu2017encoder,zhai2017neural} 
and external memory \cite{peng2015recurrent}.
In general, these works adopt a BiLSTM \cite{zhu2018a,zhu2017what} as the major labeling architecture
to extract various features, 
then use a CRF layer \cite{huang2015bidirectional} to model 
the label dependency.
We also adopt a BiLSTM-CRF model as baseline and claim that
a multi-task learning framework is working better than directly 
applying it on Chinese E-commerce dataset.
Previous works
only apply joint model of slot filling and intent detection \cite{zhang2016joint,liu2016joint}.
Our work is the first to propose a multi-task sequence labeling model with novel cascade and residual connections based on 
deep neural networks to tackle real-world slot filling problem.

\textbf{Multi-task Learning (MTL)} has attracted increasing attention
in both academia and industry recently.
By jointly learning across multiple tasks \cite{caruana1998multitask}, we can
improve performance on each task and reduce the need for labeled data.
There has been several attempts of using multi-task learning on 
sequence labeling task \cite{peng2016multi,peng2016improving,yang2017transfer},
where most of these works learn all tasks at the out-most layer.
Søgaard and Goldberg \shortcite{sogaard2016deep} is the first to 
assume the existence of a hierarchy between the different tasks in a stacking BiRNN model.
Compared to these works, our DCMTL model further improves this 
idea even thorough with cascade and residual connection.
	
\section{Conclusion}
\label{sec:conclusion}
In this paper, we tackle the real-world slot filling task 
in a novel Chinese online shopping assistant system. 
We proposed a deep multi-task sequence learning framework with 
cascade and residual connection. 
Our model achieves comparable results with 
several state-of-the-art models on the common slot filling dataset ATIS. 
On our real-world Chinese E-commerce dataset ECSA, 
our proposed model DCMTL also achieves best F1 score comparing 
to several strong baselines.
DCMTL has been deployed on the online shopping assistant of a dominant
Chinese E-commerce platform.
Online testing results show that our model meets better understanding of users utterances and improves customer’s shopping experience.
Our future research may include a joint model for category classification and slot filling. 
Active learning for slot filling can also be investigated by involving human-beings interaction with our system. 

\section*{Acknowledgments}
This work as partially supported by NSFC grants 91646205 and 61373031.
We thank the anonymous reviewers for their valuable comments.

\bibliographystyle{aaai}
\bibliography{ref}

\end{document}